\documentclass[letterpaper, 10 pt, conference]{ieeeconf}  % Comment this line out if you need a4paper

\IEEEoverridecommandlockouts                              % This command is only needed if 
\usepackage[utf8]{inputenc} 
\usepackage[T1]{fontenc} 
\usepackage{hyperref}       
\usepackage{url}            
\usepackage{booktabs}      
\usepackage{amsfonts}      
\usepackage{nicefrac}      
\usepackage{microtype}    
\usepackage{algorithmic}
\usepackage[ruled,vlined]{algorithm2e}
\usepackage{amsmath}
\usepackage{graphicx}
\usepackage[font=footnotesize, labelfont=bf]{caption}

\usepackage{hyperref}
\hypersetup{colorlinks=true,linkcolor=blue,urlcolor=blue}

\DeclareMathOperator{\trace}{tr}
\DeclareMathOperator{\diag}{diag}

\usepackage{tikz}

\usepackage{amssymb}
\usepackage{pgfplots}
\usepackage{pgfmath}
\usepgfplotslibrary{patchplots}
\usetikzlibrary{patterns, positioning, arrows}

\pgfmathsetmacro\sprayRadius{.75pt}
\pgfmathsetmacro\sprayPeriod{.8cm}

\pgfdeclarepatternformonly{spray}{\pgfpoint{-\sprayRadius}{-\sprayRadius}}{\pgfpoint{1cm + \sprayRadius}{1cm + \sprayRadius}}{\pgfpoint{\sprayPeriod}{\sprayPeriod}}{
    \foreach \x/\y in {2/53,6/52,11/48,23/49,20/47,32/46,41/47,47/51,56/52,46/44,4/43,16/42,33/41,41/37,49/35,55/31,37/35,44/30,28/37,24/36,17/37,7/38,0/31,8/29,18/31,28/30,37/28,30/27,46/24,51/21,24/23,12/24,4/21,18/19,12/16,31/21,38/18,26/16,46/16,56/12,52/10,45/8,51/4,37/12,35/7,24/9,14/9,2/12,8/6,15/4,27/0,34/1,40/1} {
    \pgfpathcircle{\pgfpoint{(\x) / 10 * \sprayPeriod}{\sprayPeriod - (\y) / 10 * \sprayPeriod}}{\sprayRadius}
   }
\pgfusepath{fill}
}

\title{\Large \bf
Federated Transfer Learning for EEG Signal Classification
}

\author{
Ce Ju$^{1}$, Dashan Gao$^{2, 3}$, Ravikiran Mane$^{4}$, Ben Tan$^{1}$, Yang liu$^{1}$ and Cuntai Guan$^{4}$
\thanks{$^{1}$WeBank Co., Ltd., {\tt\small \{ceju, btan, yangliu\}@webank.com}}
\thanks{$^{2}$Hong Kong University of Science and Technology, {\tt\small dgaoaa@connect.ust.hk}}
\thanks{$^{3}$Southern University of Science and Technology.}
\thanks{$^{4}$Nanyang Technological University, {\tt\small ravikian001@e.ntu.edu.sg, CTGuan@ntu.edu.sg}}
\thanks{*This work is supported by the Joint NTU-WeBank Research Centre.}
}

\begin{document}

\maketitle
\thispagestyle{empty}
\pagestyle{empty}

%%%%%%%%%%%%%%%%%%%%%%%%%%%%%%%%%%%%%%g%%%%%%%%%%%%%%%%%%%%%%%%%%%%%%%%%%%%%%%%%%
\begin{abstract}
The success of deep learning (DL) methods in the Brain-Computer Interfaces (BCI) field for classification of electroencephalographic (EEG) recordings has been restricted by the lack of large datasets. 
Privacy concerns associated with EEG signals limit the possibility of constructing a large EEG-BCI dataset by the conglomeration of multiple small ones for jointly training machine learning models. 
Hence, in this paper, we propose a novel privacy-preserving DL architecture named federated transfer learning (FTL) for EEG classification that is based on the federated learning framework.
Working with the single-trial covariance matrix, the proposed architecture extracts common discriminative information from multi-subject EEG data with the help of domain adaptation techniques. 
We evaluate the performance of the proposed architecture on the PhysioNet dataset for 2-class motor imagery classification. While avoiding the actual data sharing, our FTL approach achieves 2\% higher classification accuracy in a subject-adaptive analysis. Also, in the absence of multi-subject data, our architecture provides 6\% better accuracy compared to other state-of-the-art DL architectures.

\end{abstract}

%%%%%%%%%%%%%%%%%%%%%%%%%%%%%%%%%%%%%%%%%%%%%%%%%%%%%%%%%%%%%%%%%%%%%%%%%%%%%%%%
\section{INTRODUCTION}
Brain-Computer Interface systems aim to identify users' intentions from brain states. BCI has a prominent potential in the medical domain but its use has been limited by moderate decoding accuracies. In the era of deep learning, the success of BCI models for classification of electroencephalographic (EEG) recordings have been restricted by the lack of large datasets. Due to the high data collection costs, EEG-BCI data is present in the form of multiple small datasets that are scattered around the globe. Moreover, due to privacy concerns, it is difficult to create a large enough dataset by combining multiple small datasets. As EEG signals reflect brain activities in numerous aspects, the potential abuse of EEG data may lead to severe privacy violations and hence acts like General Data Protection Regulation (GDPR)~\cite{regulation2016regulation} prohibit organizations from exchanging data without explicit user approval. Therefore, it is significant to conduct a joint EEG signal analysis while protecting user privacy. 
Hence, to solve this problem, we propose to use federated learning framework in healthcare~\cite{mcmahan2016communication,yang2019federated,gao2019privacy,ju2020privacy}. Federated learning is an emerging and powerful technique which enables joint training of machine learning models using data from multiple sources without the need of actual data sharing between sources. 

Existing federated learning mainly focuses on the homogeneous dataset, where the different parties share the same feature space. For EEG signal collection, even the same equipment manufacturer may develop EEG signal acquisition equipment with varying electrode numbers, position and sampling rate let alone different equipment vendors. Such device diversity further exacerbates the scarcity of training data and yields numerous distributed heterogeneous datasets in EEG classification. Heterogeneous domain adaptation by linking different feature spaces based on labels has been studied \cite{wang2011heterogeneous11}. However, existing solutions assume a presence of multiple source domains with abundant labeled instances and one target domain with limited labeled instances as well as unlabeled instances. When applied to our problem setting, existing heterogeneous domain adaptation approaches will face accuracy drop. 
The reason for the accuracy drop is that in our setting, there are numerous clients each with limited labeled data, and none of them can be a source domain in a source-target domain pair. 

In this paper, we propose a novel neural network-based covariance method in BCI for the subject-specific analysis, and adapt it to the transfer learning setting for cross-subject learning in a subject-adaptive BCI analysis based on a federated learning framework. We call our method as \emph{Federated Transfer Learning} (FTL). Specifically, the architecture of our model is of deep neural networks, whose inputs are spatial covariance matrices of EEG signals, with the domain adaptation technique for adaptive analysis. 

In summary, following are the the major contributions of this paper: 
we propose a novel DL architecture based on the EEG spatial covariance matrix.
Our approach is adapted to multi-devices for transfer learning setting based on the federated learning framework to protect data privacy. We perform subject-specific and subject-adaptive analysis on PhysioNet EEG Motor Movement/Imagery Dataset. We show that FTL can match the classification accuracy of state-of-the-art methods while avoiding the sharing of EEG data

\section{RELATED WORK}
\subsubsection{Covariance Methods in BCI}
When we process and classify EEG signals, one cutting edge technology is to directly estimate and manipulate covariance matrices of EEG signal for source extraction instead of subspaces methods, such as principal component analysis (PCA), independent component analysis (ICA) and common spatial pattern (CSP)~\cite{lotte2007review,yger2016riemannian}. The covariance methods, also called Riemannian approach, encapsulates signal energy-based information of EEG signals. 
%For example, the spatial covariance matrix is defined as follows: Let $\bold{X} \in \mathbb{R}^{E\cdot D}$ be a short-time segment trial of EEG signal, where $E$ is the number of electrodes and $D$ is the epoch durations discretized as a number of samples. The spatial covariance matrix $\bold{S}$ of segment trial $\bold{X}$ is an $E \times E$ SPD matrix given by $\bold{S}:=\frac{1}{D-1}\cdot \bold{X} \cdot \bold{X}^T$. 
Covariance methods in BCI are promising methods to enable the direct manipulation of covariance matrices, and they are superior to the classical EEG signal processing approaches based on feature extraction~\cite{congedo2013new,yger2015averaging,barachant2011multiclass}. For example, Minimum Distance to the Mean (MDM) approach is a classifier based on the geodesic distances on SPD manifolds which consists of covariance matrices~\cite{yger2015averaging}. Barachant et al. and Yger et al. purpose Log-Euclidean kernel and Stein kernel method to classify SPD matrices based on the Riemannian distance respectively~\cite{barachant2013classification,scholkopf1998prior}.

\subsubsection{Deep Learning Methods in  BCI}
Symmetric Positive-Definite (SPD) matrix is one of the popular research objects encountered in a great variety of areas, such as medical imaging~\cite{jayasumana2013kernel} and visual recognition~\cite{huang2014learning}. 
The paradigm of deep neural networks on SPD matrices has been proposed to interpolate, manipulate and classify SPD matrices in bunch of application areas~\cite{huang2017riemannian,dong2017deep,huang2018building,huang2017deep}. 
For example, the Riemannian network architecture designed by Huang and Goal that includes BiMap layer, ReEig layer and LogEig layer on SPD matrices, has outperformed the existing \emph{state-of-the-art} methods in three typical visual classification tasks~\cite{huang2017riemannian}. 
Furthermore, deep learning methods, especially the convolutional neural network (CNN) architectures, have been explored to boost classification performance~\cite{reuderink2011subject,schirrmeister2017deep,sakhavi2018learning,lawhern2018eegnet}.

\section{METHODOLOGY}
The raw data is represented in the spatial covariance matrix as the input for our FTL architecture. Specifically, let $\bold{X} \in \mathbb{R}^{E\cdot D}$ be a short-time segment trial of EEG signal, where $E$ is the number of electrodes and $D$ is the epoch durations discretized as a number of samples. The spatial covariance matrix $\bold{S}$ of segment trial $\bold{X}$ is an $E \times E$ SPD matrix given by $\bold{S}:=\frac{1}{D-1}\cdot \bold{X} \cdot \bold{X}^T$. 

Our proposed architecture consists of following 4 layers: manifold reduction layer (\ref{projection}), common embedded space (\ref{ces}), tangent projection layer (\ref{tpl}) and federated layer (\ref{Federated}). The purpose of each layer is as follows:

\begin{itemize}
\item Manifold reduction layer (M): Spatial covariance matrices are always assumed to be on the high-dimensional SPD manifolds. This layer is the linear map from the high-dimensional SPD manifold to the low-dimensional one with undetermined weights for learning. 
\item Common embedded space (C): The common space is the low-dimensional SPD manifold whose elements are reduced from each high-dimensional SPD manifolds, which is designed only for the transfer learning setting. 
\item Tangent projection layer (T): This layer is to project the matrices on SPD manifolds to its tangent space, which is a local linear approximation of the curved space. 
\item Federated layer (F): Deep neural networks are implemented in this layer. For the transfer learning setting, parameters of neural networks are updated by the federated aggregation.
\end{itemize}

Based on the above layers, we design transfer version and non-transfer version FTL architectures for subject-adaptive analysis and subject-specific analysis respectively. For the transfer version, our FTL architecture consists of M, C, T and F. For the non-transfer version, our FTL architecture only consists of M and T. The outputs of either architecture are predicted labels. The architecture of the transfer version FTL refers to Figure \ref{manifold}.

\subsection{Notation}
EEG device captures signals from $m$ subjects, and we always assume that the EEG signals have been band-pass filtered. For the $i$ th subject, his/her filtered EEG signal is separated as a sequence of trials $\{ \bold{X}_j \}_{j=1}^{N_i}$, where $N_i$ is the number of trials. The basic assumption of Riemannian approach for signal processing is that spatial covariance matrices $\{\bold{S}_j\}_{j=1}^{N_i}$ of trials $\{ \bold{X}_j \}_{j=1}^{N_i}$ are distributed on an embedded SPD submanifold $\mathcal{M}_i \hookrightarrow  \mathbb{R}^{E_i \cdot D_i}$, where "$\hookrightarrow$" represents an \emph{embedding} map in differential geometry~\cite{petersen2006riemannian,ju2020geometric}.

\subsection{Manifold Reduction Layer}\label{projection}
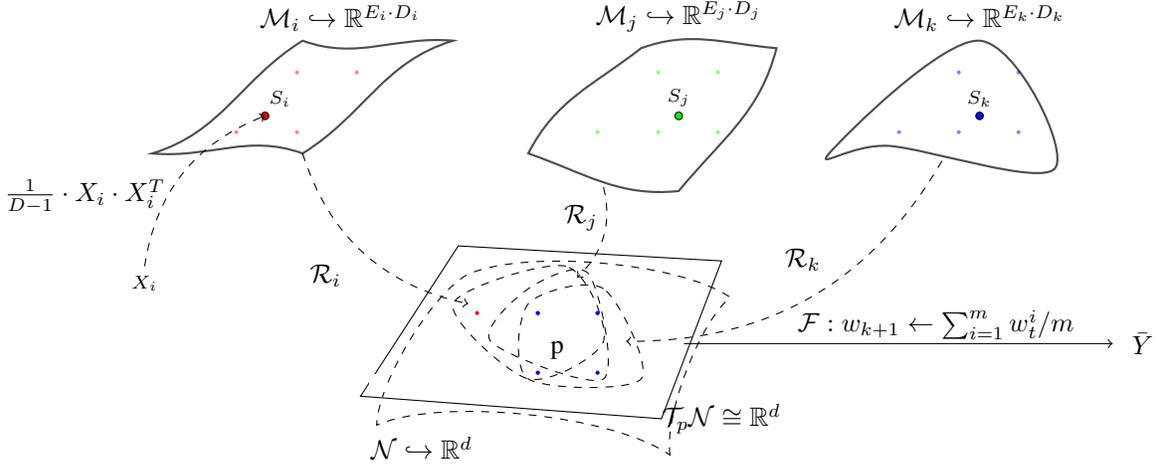
\begin{figure*}
\centering
\begin{tikzpicture}
   
    % Manifold M_1, M_2, M_3
    \begin{scope}[out=-5, in=160, relative]
    \draw [thick] (2, 1.5) to (0, 0) to (2, 0) to (4, 1.5) to cycle[smooth cycle, tension=0.5, fill=white, pattern color=red, pattern=spray, opacity=0.7]; 
    \end{scope}
     \draw [fill=red](1.5, 0.5) circle (.05); 
     \fill[black, font=\scriptsize](1.7, 0.5) node [above] {$S_i$};
     \fill[black](2.5, 1.5)  node [above] {$\mathcal{M}_i \hookrightarrow  \mathbb{R}^{E_i\cdot D_i}$};
     %Input Signal 
     \path[dashed]  [->] (-0.1, -1.5) edge [bend left] node[midway, xshift=-13mm, yshift=-3mm] {$\frac{1}{D-1} \cdot X_i \cdot X_i^T$} (1.5, 0.5);
     \fill[black, font=\scriptsize](-0.1, -1.5) node [below] {$X_i$};
     
     \begin{scope}[out=-5, in=200, relative]
     \draw [thick] (6.5, 1.4) to (5, 0) to (7, -0.5) to (8.3, 1.4) to cycle[smooth cycle, tension=0.5, fill=white, pattern color=green, pattern=spray, opacity=0.7]; 
     \end{scope}
     \draw [fill=green](7, 0.5) circle (.05); 
     \fill[black, font=\scriptsize](7, 0.5) node [above] {$S_j$};
     \fill[black](7, 1.5) node [above] {$\mathcal{M}_j \hookrightarrow  \mathbb{R}^{E_j \cdot D_j}$};
     
     \draw [thick] [smooth cycle, tension=0.6, fill=white, pattern color=blue, pattern=spray, opacity=0.7] plot coordinates{(11, 1.5) (9, 0) (9.8, 0.1) (12, -0.2)};
     \draw [fill=blue](11, 0.5) circle (.05); 
     \fill[black, font=\scriptsize](11, 0.5) node [above] {$S_k$};
      \fill[black](11, 1.5)  node [above] {$\mathcal{M}_k \hookrightarrow  \mathbb{R}^{E_k \cdot D_k}$};

    %\draw [thick] [smooth cycle, tension=0.1, fill=white, pattern color=white, pattern=north west lines, opacity=0.7] plot coordinates{(8.1, -1.5) (5, -1) (3, -2.5) (6, -3.5)} node at (7.5, -3) {$\mathcal{N} \hookrightarrow  \mathbb{R}^{d}$};
     %\draw [dashed,looseness=.6] [smooth cycle, tension=0.1, fill=white, pattern color=white, pattern=north west lines, opacity=0.7] plot coordinates{(8.1, -1.5) (5, -1) (3, -2.5) (6, -3.5)} node at (7.5, -3) {$\mathcal{N} \hookrightarrow  \mathbb{R}^{d}$};
  
  %Embedding Space N and Tangent space TpN.
  \draw (6, -4.3, -2) -- (6.8, -2.2, -2) -- (3.3, -2, -2) -- (2, -4, -2) -- cycle;
  \draw node at (6.8, -4.3, -2) {$\mathcal{T}_p \mathcal{N} \cong \mathbb{R}^{d}$};
  \draw[dashed, looseness=.6] (6.5, -4.4, -1) node at (3.2, -4.3, -1) {$\mathcal{N} \hookrightarrow  \mathbb{R}^{d}$}
  to[bend left]  (7.3, -2.3, -1)
  to[bend right] coordinate (mp) (3.8, -2, -1)
  to[bend right]  (2.6, -4, -1)
 to[bend left] coordinate (mm) (6.5, -4.4, -1) node at (5, -3, -1)  {p}
  -- cycle;
     
    %Subset of N
    \draw[dashed,smooth cycle, pattern color=red, pattern=spray] 
        plot coordinates { (6, -2.5) (5.5, -1.5) (4, -2) (5, -3) } 
        node [label={[label distance=-0.3cm, xshift=-2cm, fill=white] }] {};
        
    \draw[dashed,smooth cycle, pattern color=green, pattern=spray] 
        plot coordinates { (6, -3) (5.8, -1.6) (5, -1.8) (4.5, -2.5) } 
        node [label={[label distance=-0.3cm, xshift=-2cm, fill=white] }] {};
        
     \draw[dashed, smooth cycle, pattern color=blue, pattern=spray] 
        plot coordinates { (6.5, -3) (6, -1.8) (5, -2) (5, -3) } 
        node [label={[label distance=-0.3cm, xshift=-2cm, fill=white] }] {};

    %Function
    \path[dashed]  [->] (2, 0) edge [bend right] node[midway, xshift=-5mm, yshift=-3mm] {$\mathcal{R}_i$} (4.2, -2);
    \path[dashed] [->] (6, -0.45) edge [bend left] node[midway, xshift=-3mm, yshift=2.5mm] {$\mathcal{R}_j$} (5.65, -1.65);
    \path[dashed]  [->] (10.5, -0.1) edge [bend left] node[midway, xshift=-1mm, yshift=5mm] {$\mathcal{R}_k$} (6.3, -2.5);

   %Federated learning
    \path  [->]  (6.3, -3.3, -2) edge node[midway, xshift=5mm, yshift=2.5mm] {$\mathcal{F}: w_{k+1} \leftarrow\sum_{i=1}^m w_{t}^i /m$} (12, -3.3, -2);
    \draw node at (12.4, -3.3, -2) {$\bar{Y}$};
   
\end{tikzpicture}
\caption{Architecture of the transfer version FTL: Spatial covariance matrix $S_i \in \mathcal{M}_i$ is derived from short-time segment trial $X_i \in \mathbb{R}^{E_i \times D_i}$. In manifold reduction layer (\ref{projection}), neural networks $\mathcal{R}_i$, $\mathcal{R}_j$ and $\mathcal{R}_k$ reduce SPD manifolds $\mathcal{M}_i$, $\mathcal{M}_j$ and $\mathcal{M}_k$, respectively, on common space $\mathcal{N}$, also an SPD manifold (\ref{ces}). 
We then project the signals from $\mathcal{N}$ to tangent space $\mathcal{T}_p \mathcal{N}$ in tangent projection layer (\ref{tpl}).
Finally, we have neural networks in federated layer (\ref{Federated}) for the classification, which yields predicted labels $\bar{Y} \in \{1, \cdots, K\}$. The parameters of neural networks in this layer is updated by federated aggregation $\mathcal{F}$ in each round.
} 
\label{manifold}
\end{figure*}

For spatial covariance matrices $\{\bold{S}_1, \cdots, \bold{S}_{N_i}\}$ of the $i$ th subject on $\mathcal{M}_i \hookrightarrow  \mathbb{R}^{E_i \cdot D_i}$, we build its own reduction $\mathcal{R}_i$ to map SPD submanifold $\mathcal{M}_i $ to the common SPD manifold $\mathcal{N} \hookrightarrow  \mathbb{R}^{d}$, where $d$ is the embedding dimension of $\mathcal{N}$. Figure \ref{manifold} illustrates the manifold reduction layer. We approximate these reductions $\mathcal{R}_i$  with neural networks approach. Suppose $\bold{W}_i \in \mathbb{R}^{d \cdot E_i }$ is the weight matrix of neural network, the manifold reduction layer is constructed as a bilinear mapping, 
\[
\mathcal{R}_i (\bold{S}_j):= \bold{W}_i \cdot \bold{S}_j \cdot \bold{W}_i^T.
\]
\subsection{Common Embedded Space}\label{ces}
Upon the establishment of manifold reductions, we require that spatial covariance matrices from different subjects should fall closely onto the common embedded space $\mathcal{N}$. One way to measure the distances between two probability distributions of reduced spatial covariance matrices on common embedded space $\mathcal{N}$ is \emph{maximum mean discrepancy} (MMD) \cite{gretton2007kernel}. Suppose each projected matrices $\mathcal{R}_i \big(\{\bold{S}_1, \cdots, \bold{S}_{N_i}\} \big) \sim \mathcal{Q}_i$, where $\mathcal{Q}_i$ is the probability distribution over common manifold $\mathcal{N}$. For a feature map $\psi: \mathcal{N} \longrightarrow \mathcal{H}$, where $\mathcal{H}$ is a reproducing kernel Hilbert space (RKHS). Hence, the MMD distance between two probability distributions is as follows, 
\begin{align*}
&\text{MMD}_{\psi}(\mathcal{Q}_i, \mathcal{Q}_j)\\
:= & || \mathbb{E}_{ \mathcal{R}_i(\bold{S}) \sim\mathcal{Q}_i} \psi( \mathcal{R}_i(\bold{S}) ) -  \mathbb{E}_{ \mathcal{R}_j(\bold{S})\sim\mathcal{Q}_j} \psi( \mathcal{R}_j(\bold{S}) ) ||_{\mathcal{H}},
\end{align*}
where random variable $\bold{S} \in \{\bold{S}_1, \cdots, \bold{S}_{N_i}\}$ $(i= 1,\cdots, m)$. Notice that the small MMD yields closed reduced matrices on $\mathcal{N}$.

%%%%%%%%%%%%%%%%%%%%%%%%%%%%%%%%%%%%%%%%%%%%%%%%%%%%%%%%%%%%%%%%%%%%%%%%%%%%%%
\subsection{Tangent Projection Layer}\label{tpl}
For any reference matrix $\bold{P}$ on common embedded SPD manifold $\mathcal{N}$, we consider the associated tangent space $\mathcal{T}_{\bold{P}} \mathcal{N}$. Let $\bold{V}_1$ and $\bold{V}_2$ be two tangent vectors on tangent space $\mathcal{T}_{\bold{P}} \mathcal{N}$ with the inner product defined as $\langle \bold{V}_1, \bold{V}_2 \rangle_{\bold{P}} := \trace \big( \bold{V}_1\cdot \bold{P}^{-1} \cdot \bold{V}_2 \cdot \bold{P}^{-1} \big)$. Logarithmic map LOG on manifolds locally project the spatial covariance matrix $\tilde{\bold{P}}$ onto the tangent space of reference matrix $\bold{P}$ by
\begin{align*}
\bold{V} :&= \text{LOG}_{\bold{P}} (\tilde{\bold{P}}) \\
               &= \bold{P}^{\frac{1}{2}} \cdot \log \big( \bold{P}^{-\frac{1}{2}} \cdot \tilde{\bold{P}} \cdot \bold{P}^{-\frac{1}{2}} \big) \cdot  \bold{P}^{\frac{1}{2}},
\end{align*}
where $\log$ is the logarithm of SPD matrix $\bold{P} = \bold{U} \cdot \diag \big( \sigma_1, \cdots, \sigma_E 
\big) \cdot \bold{U}^T$, such that
\[
\log (\bold{P}) := \bold{U} \cdot \diag \big( \log(\sigma_1), \cdots, \log(\sigma_E) \big) \cdot \bold{U}^T.
\]

Specifically, the reference matrix $\bold{P} = \bold{I}_E$ yields a compact expression of tangent vector $\bold{V} = \log (\tilde{\bold{P}})$. Hence, we let the tangent projection layer $\mathcal{P}_{\mathcal{N}}$ on common space $\mathcal{N}$ be $\mathcal{P}_{\mathcal{N}}(\bold{X}):= \log (\bold{X})$, where $\bold{X}$ is the set of all reduced spatial covariance matrices from different subjects on $\mathcal{N}$.

%%%%%%%%%%%%%%%%%%%%%%%%%%%%%%%%%%%%%%%%%
\subsection{Federated layer}\label{Federated}
The tangent space of common manifold $\mathcal{N}$ is locally homeomorphic in the Euclidean space of dimension $d$. Hence, we establish typical neural networks as the classifier from tangent space. The architectures of each subject's classifier in federated layer are assumed to be the same including fully connected layer, flatten layer and activation function. 

At training round $t$, we adopt federated averaging method\cite{DBLP:journals/corr/McMahanMRA16} to train the model weights $w_t$ in federated layer. Federated averaging is conducted over clients of each subject for feature map aggregation and over all the clients for classifier aggregation. Specifically, the techniques of federated learning follows a server-client setting. A server acts as model aggregator. In each round, the server collects updated local models from each client for model aggregation. After model aggregation, the server sends the updated global model to each client as follows,
\[
\mathcal{F}: w_{t+1} \leftarrow \frac{1}{m}\cdot \sum_{i=1}^m w_{t}^i
\] 

When a client receives the model sent by server, it updates the model with its local data distributed on the reduced common manifold $\mathcal{R}_i (\mathcal{M}_i) \subset \mathcal{N} \hookrightarrow  \mathbb{R}^{d}$. The training process continues until the model converges. 

\subsection{Architecture and Loss}
There are two classes of loss in our approach including the classification loss and the domain loss as follows, 
\begin{itemize}
\item Classification Loss: the cross entropy loss between predicted label and ground truth, write as $\mathcal{L}_C(\bar{Y}, Y)$, where $\bar{Y}, Y \in \{1,\cdots, K\}$ is predicted label and ground truth respectively.
\item Domain Loss: MMD$_{\psi}(\mathcal{Q}_i^Y, \mathcal{Q}_j^Y)$ is MMD distance with pre-set feature map $\psi$ between any two of probability distributions $\mathcal{Q}_i^Y$ and $\mathcal{Q}_j^Y$, for $1\leq i < j \leq m$ and $Y  \in \{1,\cdots, K\}$. 
\end{itemize}

For the non-transfer version, the architecture includes repeated manifold reduction layer, tangent projection layer and typical neural networks with only the classification loss, i.e.
\begin{align*}
\mathcal{L} := \sum_{i=1}^m \mathcal{L}_C(\bar{Y}_i, Y),
\end{align*}
where subscript $i$ represents each subject. 

For the transfer version, the architecture includes two classes of losses as follows, 
\begin{align*}
\mathcal{L} := \sum_{i=1}^m \mathcal{L}_C(\bar{Y}_i, Y) + \sum_{1 \leq i < j \leq m} \sum_{Y=1}^K \lambda_{i, j}^Y \cdot \text{MMD}_{\psi}(\mathcal{Q}_i^Y, \mathcal{Q}_j^Y),
\end{align*}
where $\lambda_{i, j}^Y$ are pre-set weights for each domain loss term. 

\begin{figure*}
\centering
	\includegraphics[height = 0.18\textheight, width=\textwidth]{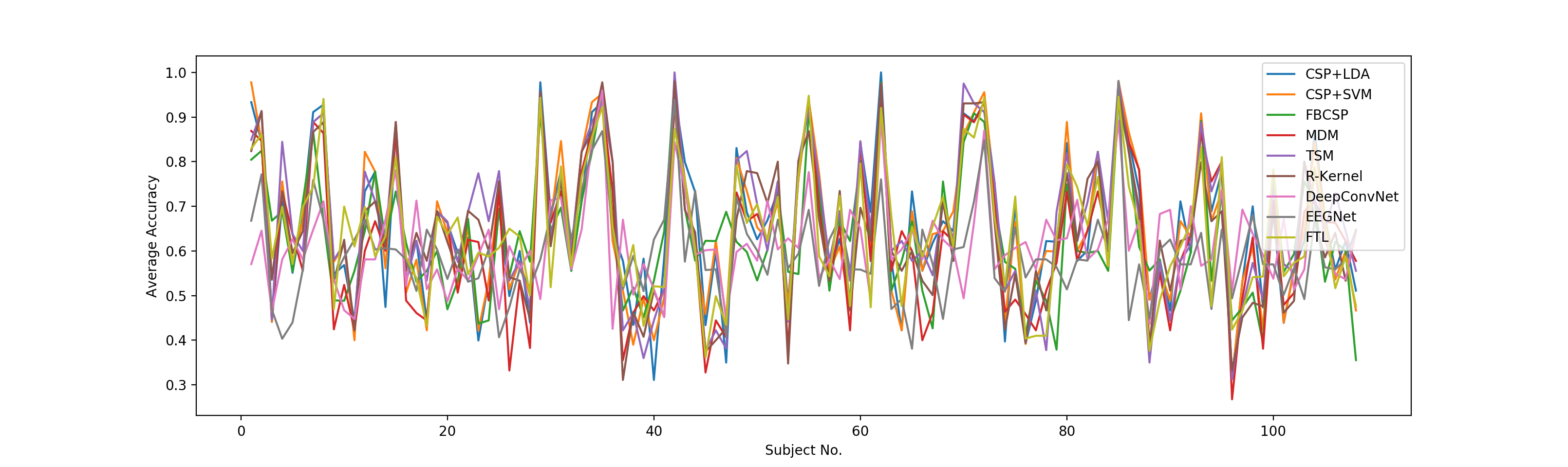}
	\caption{Average accuracy of classifiers on 109 subject in subject-specific analysis: horizontal axis represents the subject number from 1 to 109, and vertical axis represents the average accuracy.}
	\label{average}
\end{figure*}

\section{EXPERIMENTS}
\subsection{Experimental Setup}
The EEG data used in our experiments is from PhysioNet EEG Motor Imagery(MI) Dataset~\cite{schalk2004bci2000}. 
\subsubsection{MI Dataset Description}
The PhysioNet EEG dataset is recorded from 109 subjects. In our experiments, we consider the 2-class MI data, i.e. imagined left and right hand movements. The EEG data is recorded using 64 electrodes at 160Hz sampling frequency. 

\subsubsection{Evaluated Algorithms}\label{EA}
We evaluated FTL against the following subspaces methods, covariance methods and DL methods:
\begin{itemize}
\item CSP + linear discriminant analysis (LDA)/support vector machines (SVM): CSP spatial filtering algorithm with an LDA/SVM classifier~\cite{lotte2007review,barachant2011multiclass}.
\item FBCSP: Best-performing CSP method for motor imagery classification~\cite{ang2008filter,chin2009multi}.
\item MDM: Classification algorithm based on computing the geodesic distances on SPD manifolds ~\cite{barachant2011multiclass}.
\item Tangent Space Mapping (TSM): Classification algorithm on the tangent space of SPD manifolds~\cite{barachant2011multiclass}.
\item Riemannian-based Kernel Method (R-Kernal): Kernel method with the specific kernel derived from Riemannian geometry of SPD matrices~\cite{barachant2013classification}. 
\item EEGNet: A compact CNN architecture for EEG-based BCIs~\cite{lawhern2018eegnet}. 
\item DeepConvNet: A new discriminative spectral–spatial input to represent a diversity of brain signal patterns across the subjects and sessions~\cite{kwon2019subject}.
\end{itemize} 

\subsubsection{Architecture of FTL in Experiments}
All the experiments are conducted on a machine with 2.6 GHz 6-Core Intel Core i7 with Memory 16 GB 2400 MHz DDR4.
\begin{itemize}
\item Subject-specific Analysis: The architecture of FTL in this analysis includes three-manifold reduction layers (i.e. $\bold{W}_1 \in \mathbb{R}^{32 \times 4}$, $\bold{W}_2 \in \mathbb{R}^{4 \times 4}$ and $\bold{W}_3 \in \mathbb{R}^{4 \times 4}$), tangent projection layer and federated layer without federated aggregation. The learning rate is initialized to 0.1 with a decay after 50 epochs. We stop the training procedure when the training loss is less than 0.1. 
\item Subject-adaptive Analysis: The architecture of FTL in this analysis includes two clients, i.e. source domain (good subjects) and target domain (bad subjects). For the source domain (27 good subjects), we design three-manifold reduction layers (i.e. $\bold{W}_1 \in \mathbb{R}^{32 \times16}$, $\bold{W}_2 \in \mathbb{R}^{16 \times 8}$ and $\bold{W}_3 \in \mathbb{R}^{8 \times 4}$). For the target domain (1 bad subject), we design a small two-manifold reduction layers (i.e. $\bold{W}_1 \in \mathbb{R}^{32 \times 8}$ and $\bold{W}_2 \in \mathbb{R}^{8 \times 4}$). According to our construction, common space $\mathcal{N}$ is a SPD manifold with dimension $4\times 4$. The kernel in RKHS of $\mathcal{N}$ is Gaussian kernel with a mean of 0 and a population standard deviation 2. The federated layer is a simple fully-connected structure $\bold{W} \in \mathbb{R}^{16 \times 2}$ with federated aggregation. In the loss, the pre-set weights are as follows: $\lambda_{1,2}^1 = \lambda_{1,2}^2 = 0.1$. Additionally, the learning rates are initialized to 0.1 with a 2\% decay after 50 epochs. We stop the training procedure when the training loss is less than 1.5. 
\end{itemize}

\begin{table}
\begin{center}
\begin{tabular}{| l | l | l | l | l |  }
\hline
\emph{Subspace Methods}&  CSP+LDA & CSP+SVM & FBCSP  \\ \hline
\emph{Avg Acc.}& 0.654 &  0.653    &  0.631    \\ \hline \hline
\emph{Covariance Methods}  &MDM &TSM & R-Kernel \\ \hline
\emph{Avg Acc.}  &0.627 & \bf{0.663}  & 0.644  \\ \hline \hline
\emph{DL Methods} & EEGNet &DeepConvNet &FTL\\ \hline
\emph{Avg Acc.}  & 0.567  & 0.574  & 0.633\\ \hline
\end{tabular}
\end{center}
\caption{Cross-validation average accuracy (Avg Acc.) of 9 classifiers on 109 subjects. We have 3 subspaces methods, 3 covariance methods and 3 DL methods in the table.\label{table1}}
\end{table}

\begin{table}
\begin{center}
\begin{tabular}{| l | l  | l | l | l}
\hline
 Alg. /Setting &\emph{Bad} Subject &\emph{Good} Subject & Transfer Learning \\ \hline
 CSP+LDA&0.489  & 0.869  & 0.528   \\ \hline
 CSP+SVM& 0.488 & \bf{0.876} &  0.527 \\ \hline
 FBCSP   & \bf{0.532} &  0.805   &  0.525     \\ \hline \hline
 MDM & 0.478 & 0.842 & 0.522  \\ \hline
 TSM & 0.519 & 0.868 &  0.547 \\ \hline
 R-Kernel & 0.514 & 0.840 &  0.537 \\ \hline \hline
 EEGNet &0.487 &0.719 & 0.513\\ \hline 
 DeepConvNet &0.505&0.721& 0.520\\ \hline
 FTL & 0.513 & 0.837 & \bf{0.549}\\ \hline
\end{tabular}
\end{center}
\caption{Cross-validation accuracy of 9 classifiers on three experimental settings. The results in first and second columns are average accuracies of 27 \emph{good} subjects and 28 \emph{bad} subjects derived from the subject-specific analysis respectively. 
\label{table2}}
\end{table}

\subsection{Experimental Results and Analysis}
Our experimental section includes a subject-specific analysis and a subject-adaptive analysis. 
All experiments are evaluated in 5-fold cross-validation settings with 4 folds being used for training and 1 fold for testing. Trials were randomly assigned to different folds and this allocation was maintained constant across the classification methods.
The result is plotted in Figure \ref{average}, and it demonstrates that no competing classifier outperforms others. The specific experimental setting is as follows, 

\subsubsection{Subject-specific Analysis for 2-class MI Classification} 
Table \ref{table1} demonstrates that the average performance for each classifiers of 109 subjects is around 0.60. It is worthy to mention that subspace methods and covariance methods usually perform better than DL methods on a small sample size dataset. However, our approach FTL takes advantage of the covariance methods, and outperforms other two state-of-the-art DL methods in this experiment. 

We pick \emph{good} and \emph{bad} subjects from 109 subjects for the following subject adaptive analysis, where the word \emph{good} and \emph{bad} are in the following sense that the classification accuracy of CSP+LDA classifiers is in Top $25\%$ and Bottom $25\%$ of 109 subjects respectively. According to this construction, we have 27 \emph{good} subjects and 28 \emph{bad} subjects.

\subsubsection{Subject-adaptive Analysis for 2-class MI Classification}
In subject-adaptive analysis experiment, we have three experimental settings, i.e. \emph{bad} subject setting, \emph{good} subject setting and transfer learning setting. \emph{Bad} and \emph{good} subject settings refers to the average of results in subject-specific analysis for \emph{bad} and \emph{good} subjects respectively.
%in which the training data for every bad subject is 4 folds of the bad subjects data and the testing data is the rest 1 fold. 
%
Transfer learning setting refers to the subject-adaptive analysis in BCI that the training data for every bad subject is good subject data combined with 4 folds of the bad subjects data, and the testing data is the rest 1 fold.  
The results of the transfer learning setting, listed in the third column of Table \ref{table2}, demonstrates that modeling with signals from good subjects yields better accuracy than modeling only with the bad subject (the results in the first column). Furthermore, FTL is an effective classifier, which is equipped with the techniques of domain adaptation, for transfer learning setting.

\subsection{Discussion}
Lack of adequate training data is a major challenge in the adaption of DL methods for EEG-BCI classification. In this paper, to mitigate this issue, we proposed a federated transfer learning based deep learning architectures for subject-adaptive EEG-MI classification. 
Furthermore, following the successful Riemannian approach~\cite{yger2016riemannian,congedo2013new,yger2015averaging,barachant2011multiclass,barachant2013classification}, our architecture used signal covariance matrix as an input. 
With this architectures, in a subject-specific analysis, we achieved 6\% better classification accuracy compared to the other two state-of-the-art DL architectures. This indicates that the covariance-based representation of EEG can be an effective input for DL architectures, particularly in the absence of adequate training data.
Furthermore, in a subject-adaptive analysis, the proposed method achieved the best classification accuracy which shows that the use of domain adaptation in the FTL architecture can boost classification performance. 
Lastly, as done in this work, the federated learning framework can be effectively used to enable distributed training of EEG classifiers from multiple heterogeneous data configurations.

\section{CONCLUSIONS}
In this paper, we investigated the feasibility of the federated learning framework to enable a distributed training of BCI models from multiple datasets with heterogeneous configurations. The experimental results confirm the effectiveness of our approach and demonstrate the prominent potential of extracting knowledge from the heterogeneous BCI data.

\section{AKNOWLEDGEMENT}
We would like to thank WeBank FATE developer community, Tianjian Chen (WeBank Co., Ltd.), Yuan Jin (Shenzhen Gradient Technology Co., Ltd.) and Ruihui Zhao (Tencent Jarvis Lab) for their useful suggestions and contributions. 

{\footnotesize
\bibliographystyle{unsrt}
\bibliography{refs}
}

\end{document}